\documentclass[final]{cvpr}

\usepackage{times}
\usepackage{epsfig}
\usepackage{graphicx}
\usepackage{amsmath}
\usepackage{amssymb}
\usepackage{verbatim}
\graphicspath{{images/}}
\usepackage{subcaption}

\usepackage[pagebackref=true,breaklinks=true,colorlinks,bookmarks=false]{hyperref}

\usepackage{diagbox}

\makeatletter
\robustify\@latex@warning@no@line
\makeatother
\usepackage{authblk}

\def\confYear{CVPR 2021}
\begin{document}
\title{Can self-training identify suspicious ugly duckling lesions?}

\author[1, 2]{Mohammadreza Mohseni\thanks{Corresponding Author}}
\author[2]{Jordan Yap}
\author[2]{William Yolland}
\author[4]{Arash Koochek}
\author[1, 3]{M Stella Atkins}
\affil[1]{School of Computing Science, Simon Fraser University}
\affil[2]{MetaOptima Technology Inc}
\affil[3]{Department of Skin Science and Dermatology, University of British Columbia}
\affil[4]{Banner Health}
\affil[ ] {\vspace{-3mm}}
\affil[ ]{ \tt\small\{mmohseni, stella\}@sfu.ca, \{jordan, william\}@metaoptima.com, arash.koochek@bannerhealth.com}

\maketitle

\begin{abstract}
   One commonly used clinical approach towards detecting melanomas recognises the existence of Ugly Duckling nevi, or skin lesions which look different from the other lesions on the same patient. An automatic method of detecting and analysing these lesions would help to standardize studies, compared with manual screening methods. However, it is difficult to obtain expertly-labelled images for ugly duckling lesions. We therefore propose to use self-supervised machine learning to automatically detect outlier lesions. We first automatically detect and extract all the lesions from a wide-field skin image, and calculate an embedding for each detected lesion in a patient image, based on automatically identified features. These embeddings are then used to calculate the L2 distances as a way to measure dissimilarity. Using this deep learning method, Ugly Ducklings are identified as outliers which should deserve more attention from the examining physician. We evaluate through comparison with dermatologists, and achieve a sensitivity rate of 72.1\% and diagnostic accuracy of 94.2\% on the held-out test set.
\end{abstract}

\section{Introduction}

In order to diagnose skin cancers such as malignant melanoma, careful visual inspection of all the patient’s skin lesions, particularly melanocytic nevi, must be undertaken. To this end, dermatologists have introduced several approaches which can be helpful in detecting melanomas, including the well-known ABCD dermatology criteria (Asymmetry, Border irregularity, Color irregularity, Diameter$>$6mm)\cite{ABCDNachbar, ABCDAbbasi} and 7-point checklist\cite{SevenPointArgenziano, SevenPointBetta}. 
Another useful clinical indicator is the existence of Ugly Duckling (UD) lesions, which look different from the other lesions on the same patient\cite{GrobUD}. The existence of a UD lesion is strongly correlated with the existence of melanoma \cite{GaudyMarqesteUD, ScopeUDLetter}, although there is considerable inter-observer variability in visually selecting UD lesions from whole body imaging \cite{IlyasUD, ScopeUD}.
After identification of UDs, clinically concerning lesions can then be inspected with a close-up examination, for example under dermatoscopic inspection.
However, converting the clinical view of a small 1.5mm diameter nevus into a suitable digital representation for imaging and automatically identifying suspicious lesions is challenging.

Automatic UD detection can be treated as a form of outlier detection problem whose data comes from wide-field (clinical) images of the same patient, typically acquired during total body photography (TBP) also known as full body imaging. In TBP, overview images are typically taken with cameras placed 20-50 cm away from the skin surface, showing enough context to identify the body part, mimicking the visual inspection made by a clinician. 

However a limitation of UD analysis from TBP images is lesion size. With current camera resolutions, very small lesions less than 1.5mm diameter cannot be readily detected in a TBP image of one body part. And just like a physical examination, a closer-up view is needed for lesion diagnosis.

\begin{figure}[htp]
    \centering
    \includegraphics[width=8cm]{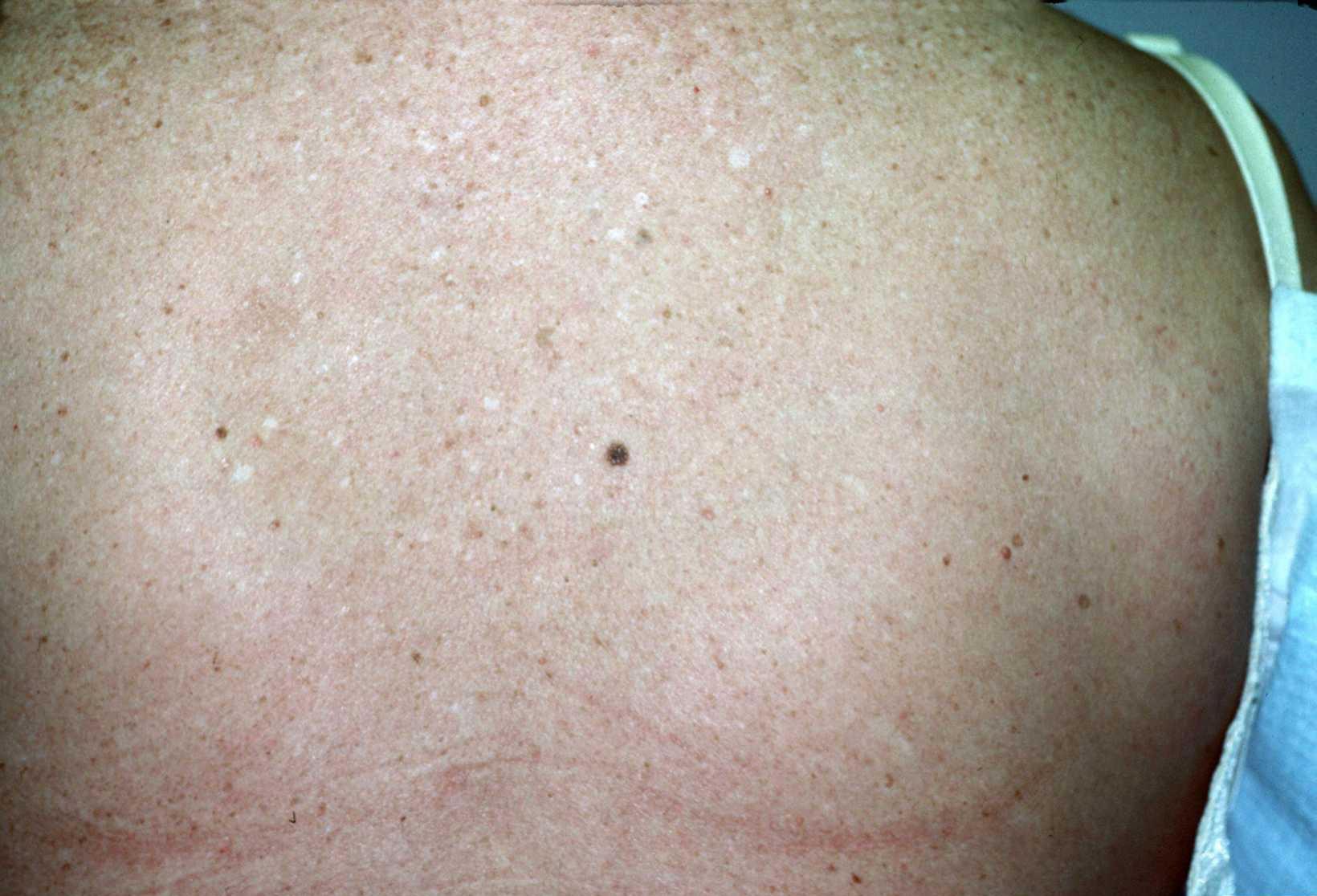}
    \caption{Typical image of part of a back, 1640 x 1116 pixels, used with permission from the SD-260 dataset \cite{SD-260}}
    \label{fig:backHB}
\end{figure}

An  example of a typical TBP image taken with a smartphone camera from about 50cm away is shown in Fig \ref{fig:backHB}. The filesize after this image has been compressed using JPEG is 187 KB.

Converting the clinical view of a small 1.5mm diameter nevus into a suitable digital representation for imaging and diagnosing suspicious nevi is challenging. In our experience, the lesion should occupy an area of at least 20x20 pixels to be reliably detected by algorithms and for automated analysis. Therefore, a small 1.5mm diameter lesion requires resolution of around 0.075mm/pixel at the imaging surface.  A wide-field digital image taken at about 20cm from the subject with a 10 MP smartphone camera, allows digital resolution of around 0.075mm/pixel \cite{dugonik2020};  smaller lesions cannot be reliably detected or analysed \cite{SoenksenUD2021}.
 Artifacts from reflections, lack of polarization, image compression, motion blurring, lighting, and shadows all contribute to the challenge of both identifying UDs from a single digital wide-field image, as well as in detecting changes in lesions found in a series of TBP images taken over time. 

\begin{figure}[htp]
\centering
\begin{subfigure}{0.45\linewidth}
\includegraphics[width=0.9\linewidth]{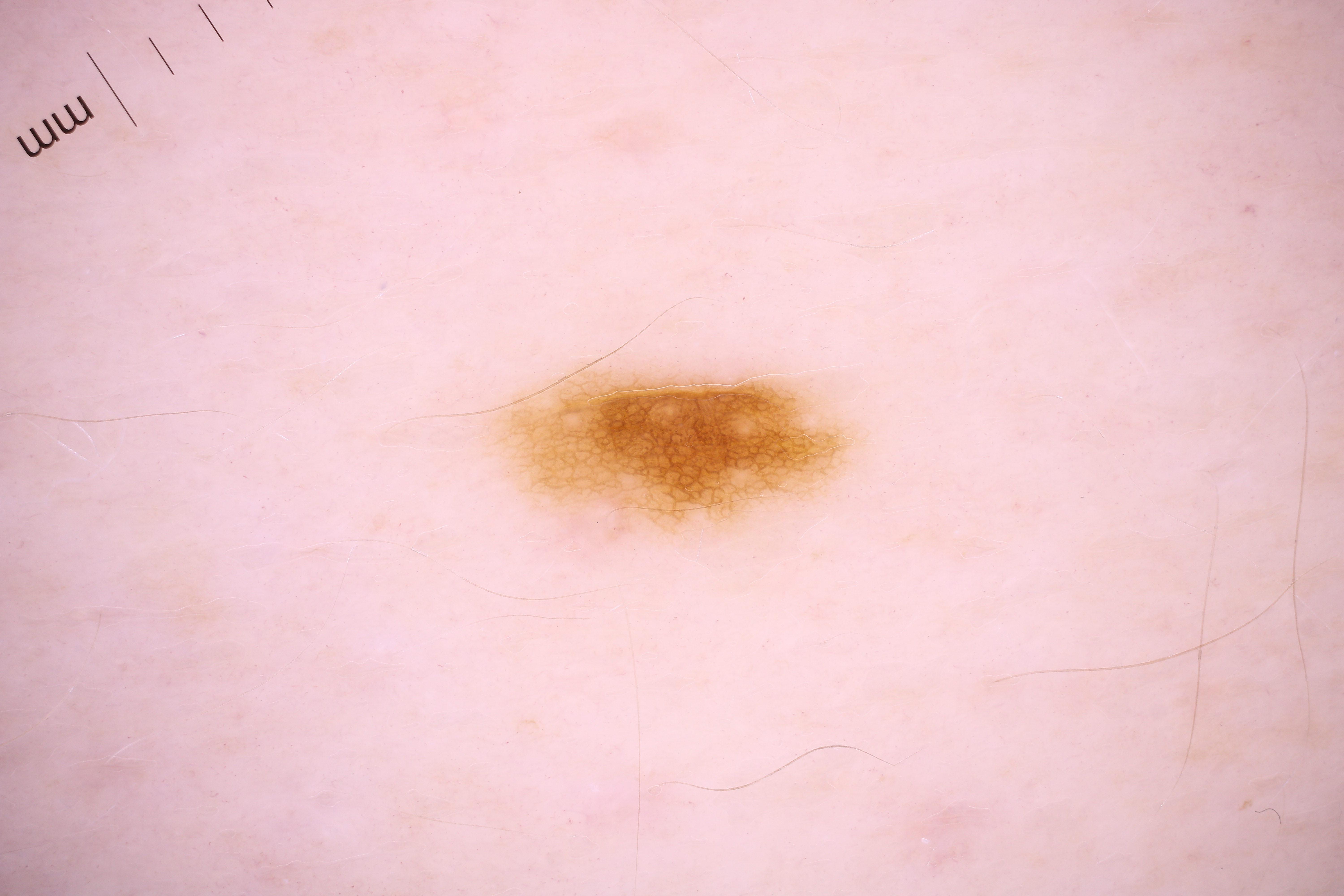} 
\caption{Dermoscopic nevus}
\label{fig:dermnevus}
\end{subfigure}
\begin{subfigure}{0.45\linewidth}
\includegraphics[width=0.9\linewidth]{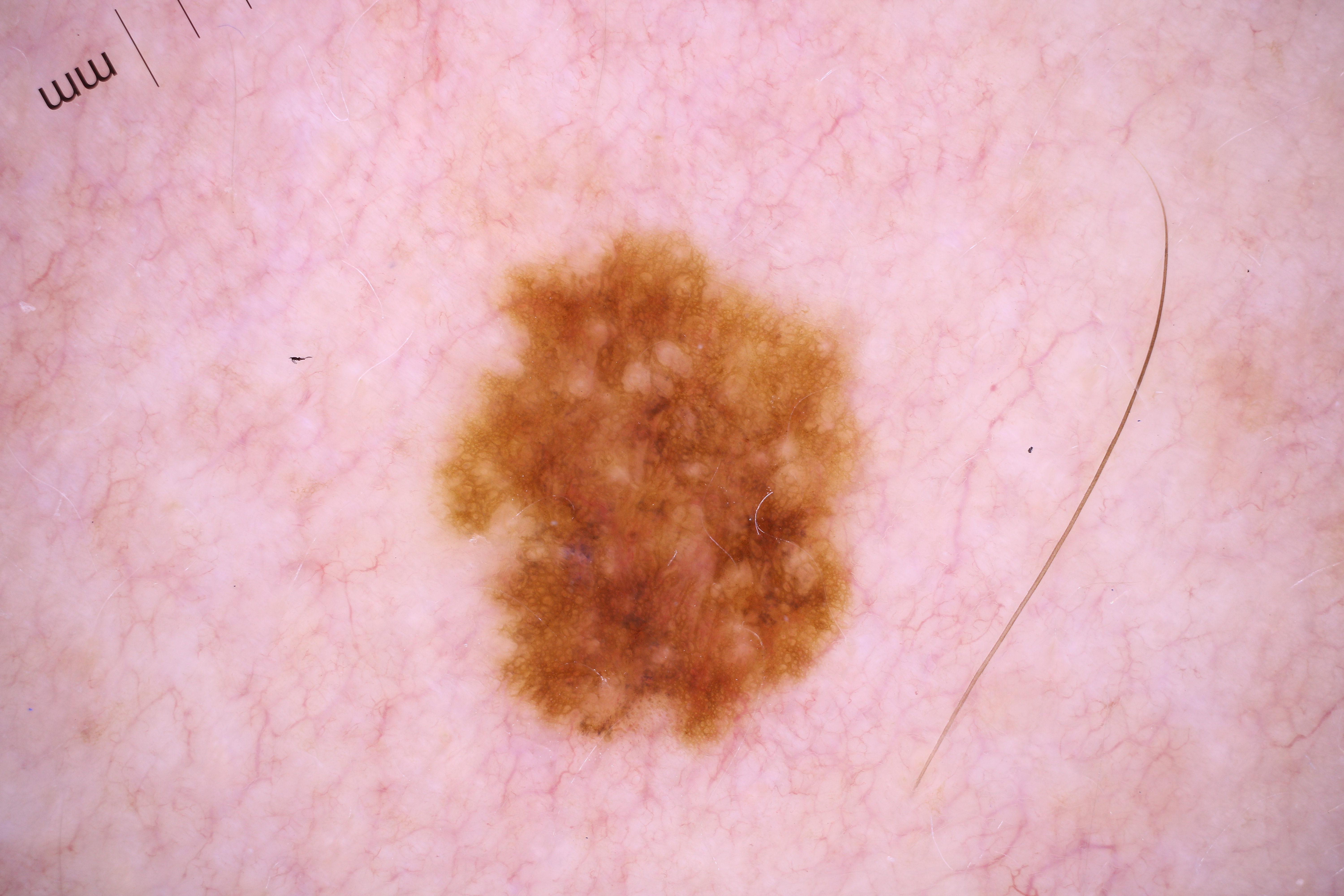}
\caption{Dermoscopic melanoma}
\label{fig:dermmel}
\end{subfigure}

\caption{Two dermoscopic images from the same patient, each of size 6000x4000 pixels (publicly available skin images from ISIC \cite{HAM_Dataset} ). Note the mm scale ruler in the top left corner of each image.}
\label{fig:derm}
\end{figure}

Difficulties in obtaining high enough resolution TBP images from personal cameras such as smartphones for automated analysis of small lesions, has also led to difficulties in obtaining expertly-labelled digital TBP images. In our experience, the expert reader cannot zoom the image enough to make an analysis, unlike in a clinical environment where the physician can look closer at suspicious lesions using a magnifying dermoscope. For example, dermoscopic images are shown in Fig \ref{fig:derm}.The size of these dermoscopic images are 6000 x 4000 pixels, JPEG Size 1 MB, providing much more detail for machine learning classification algorithms. 

Even very small lesions such as in Fig \ref{fig:dermnevus}, where the nevus is about 3mm wide, occupy about 30,000 pixels.

A few commercial systems can provide automated analysis of digital TBP images, where the images are taken with several fixed cameras for super-high resolution. One is the Canfield 3D Vectra system, where the lesion images are automatically stitched together, and then suspicious lesions can be viewed clinically with a dermoscope \cite{Canfield}. Another system which supports TBP using mobile cameras is provided by Dermengine 
\cite{dermengine}. An overview of such systems is provided by the International Society for Digital Imaging of the Skin (ISDIS) [ISDIS 2021
\cite{ISDIS}].


Automated analysis of TBP "wide-field" digital images taken with smartphones remains challenging, because of the difficulty of classifying and labelling such small lesions in the images.
We hypothesised that outlier and suspicious UD lesions could be identified using a self-supervised learning method, leading to identification of suspicious lesions which can then be chosen for further detailed analysis (by computer algorithm and/or examination by a board-certified dermatologist).

In this paper we describe our objective, unbiased, and reproducible method to automatically identify UDs in TBP images, achieving accuracy comparable to that of expert dermatologists. By using our approach, expert dermatologists can be helped to identify which lesions deserve further examination in patients with numerous atypical lesions. These tasks are repetitive and time-consuming, with poor intra-observer and inter-observer reliability and consistency. 
Furthermore, as Schlessinger \etal suggested \cite{Schlessinger2019}, our method can also be used in objective tasks needed for various outcome measures in clinical trials involving TBP images.

\begin{figure*}[htp]
    \centering
    \includegraphics[width=0.9\linewidth]{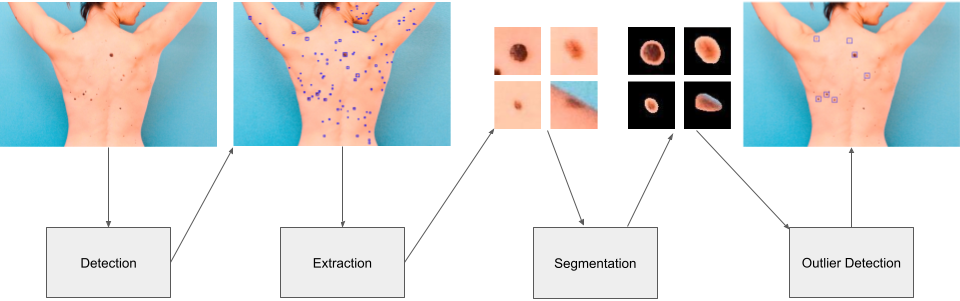}
    \caption{UD Detection Pipeline}
    \label{fig:UD_Pipeline}
\end{figure*}

\section{Related Work}

With the recent advances in deep learning and its promising results, deep learning found its way to various areas of science and dermatology was no exception. Esteva \etal showed how promising deep learning can be in skin cancer classification \cite{Esteva2017}. Liu \etal extended this success to classification of a large number of general dermatology and skin cancer classes \cite{Liu_GeneralDerm_2020}. The contribution of deep learning has not been limited to classification. It has also contributed to segmentation and detection tasks.\\

One still unsolved challenge in deep learning is outlier detection. There have been many solutions proposed to overcome this challenge but no final solution is known yet. In the literature, this problem is known by different names such as outlier detection, anomaly detection, or one-class classification. A large body of research has demonstrated that finding a good lower dimension representation can lead to promising solutions to outlier detection problems \cite{Schlegl_AnoGAN_2017, chalapathy_robust_2017, aggarwal_outlier_2017}. Perera \etal showed that representations learned for one-class classification, can be transferred from an unrelated task\cite{perera_learning_2019}. Autoencoders are a family of neural networks which learn low-dimensional representation from the input domain. Variational autoencoders (VAE) are a subtype of autoencoders which control the latent space by enforcing a prior \cite{kingma2013vae}. Higgins \etal showed that VAEs are able to learn disentangled features\cite{higgins2016early} and introduced parameter $\beta$ and $\beta$-VAE as a means for disentangled representation learning\cite{higgins2016beta}.\\

Despite the recent advances in machine learning and deep learning, UD lesion detection still remains a challenge in computer aided dermatology. Soenksen \etal recently published a supervised machine learning method to identify UD  lesions using the features of a pretrained network \cite{SoenksenUD2021, birkenfeld2020}. Three board-certified dermatologists labelled UDs by rank order in 135 TBP patient images. Their results show between 83\% and 88\%  agreement between the top 3 UDs ranked by the dermatologists and the top 3 outlier lesions ranked by the algorithm.  

In this paper we show that it is possible to skip the long process of collecting labelled data and training a classifier. We propose self-training as an approach which enables researchers to reproduce our results easily; and comparably does as good job as using the labelled data.

\section{Method}
In order to detect and classify each lesion as an inlier or UD, we developed a multi-stage pipeline which consists of three main modules (shown in Fig \ref{fig:UD_Pipeline}). First, we pass the input TBP image to the detection module. This module detects all the lesions in the TBP image, crops a fixed sized window centered on the lesion and passes them to the segmentation module. The segmentation module segments the lesions from the skin and prepares the segmented images for the outlier detection module which is the last step.

\subsection{Detection}
The detection module used was based on the Single Shot MultiBox Detector (SSD) \cite{LiuSSD2016}. We used Resnet-18 \cite{DBLP:journals/corr/HeZRS15} as our backbone along with a Feature Pyramid Network (FPN) \cite{lin2017feature} for improved multi-scale detection performance. The detection network was trained using 512x512 images as input to the network.

To detect lesions across a large TBP image, first the full-sized image is split up into 512x512 tiles with a 50\% overlap. Tiles are then sent through the detection module to obtain the predicted locations of lesions. Detections are then aggregated across all tiles and non-max suppression is performed to obtain the final set of lesion detections for the entire TBP image.

\subsection{Segmentation}

The segmentation module is a smaller variant of the U-Net architecture\cite{RonnebergerUNet2015}, with only 6.2\% of the trainable parameters as the original model. We use no batch normalization\cite{ioffe2015batch} in the model, similar to the original paper, and unlike some of the recent implementations in deep learning libraries\cite{NEURIPS2019_9015}. The model runs on 64x64 RGB images, cropped around detected lesions, and produces a 64x64 probability map. An optimal threshold is determined on a validation set for binary lesion/background segmentation.

The model is trained on thousands of manually segmented lesions along with a collection of skin-only patches all of which are sourced from a set of internal full-body images spanning various skin-types, and body locations. We found that without the true negative patches, the model produced false positive segmentation masks when presented with detection errors (skin-only patches).  We train with random rotations, flips, crops, and color-jitter, and optimize for a pixel-wise cross-entropy loss using the Adam\cite{kingma2014adam} optimizer with a fixed step-down schedule and an initial learning rate of $1e^{-3}$.

\subsection{Outlier Detection}
Variational autoencoders tend to have higher reconstruction loss on anomalous samples\cite{an2015variational, VAE_Skin_Anomaly, liu_towards_vae}. This fact suggests that by training a VAE on the extracted lesions from one patient, UDs end up with high reconstruction loss value. The intuition behind this statement is that VAEs try to reconstruct common looking lesions (the majority) as perfectly as possible. UD lesions (which are the minority) are less prioritized, thus leading to higher reconstruction losses.\\

\begin{figure}[htp]
    \centering
    \includegraphics[width=8cm]{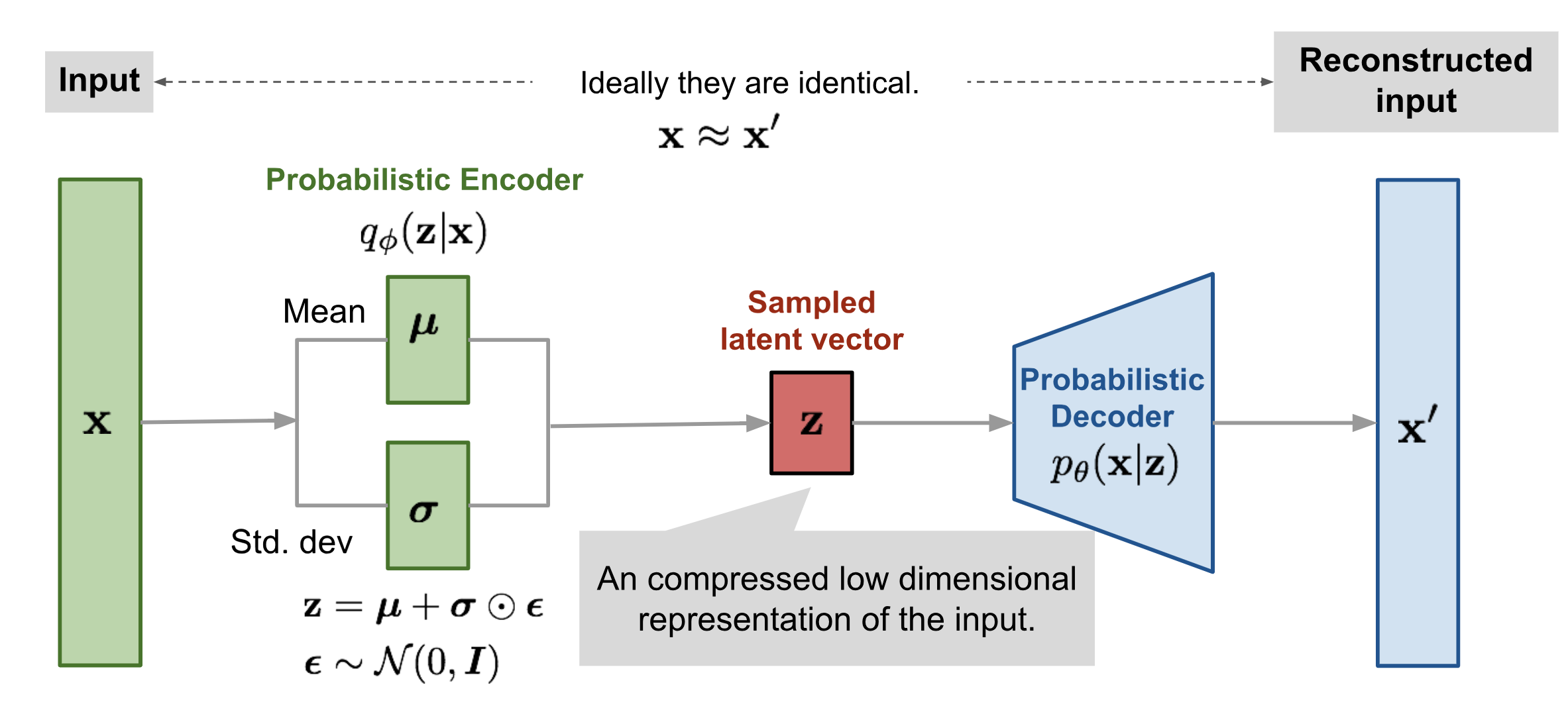}
    \caption{An overview of a Variational Autoencoder (VAE) architecture (image taken from \cite{weng2018VAE})}
    \label{fig:vae}
\end{figure}

In our first experiments we self-trained a VAE on extracted lesion images of one patient, and by defining a threshold on the reconstructed losses, we classified each lesion as either common looking or as a UD. In order to train the VAE, we enforce it to reconstruct the input image in the output as shown in Fig \ref{fig:vae}.\\

This approach's limitation is that for each TBP image query, the VAE needs to be trained for almost 130 epochs which takes $\approx 2$ minutes. We tried to overcome this issue by doing pre-training to speed things up.\\

\subsection{Fast Outlier Detection}
\begin{figure}[htp]
\centering
\begin{subfigure}{0.49\linewidth}
\includegraphics[width=\linewidth]{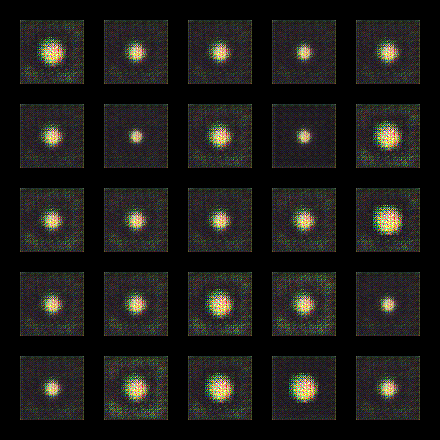} 
\caption{Reconstructed lesion images at Epoch 13}
\label{fig:epoch13}
\end{subfigure}
\begin{subfigure}{0.49\linewidth}
\includegraphics[width=\linewidth]{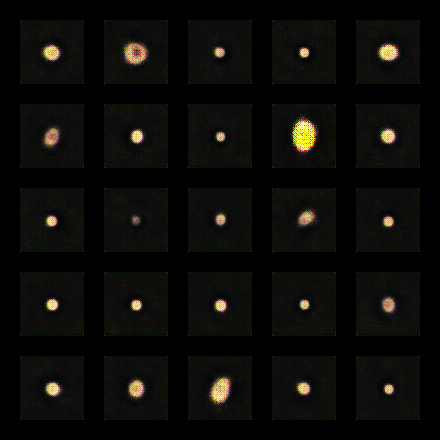}
\caption{Reconstructed lesion images at Epoch 121}
\label{fig:epoch121}
\end{subfigure}

\caption{Two unordered sets of reconstructed images at different epochs. The model has more information regarding input domain in epoch 121 compared to epoch 13.}
\label{fig:reconstructed}
\end{figure}

We observed when training the VAE from scratch, in the initial epochs VAE is learning the basics of describing the lesions as shown in Fig \ref{fig:epoch13}. With the increase in the number of epochs, our model learns more about the input domain and reconstructs better quality images in the output (as shown in Fig \ref{fig:epoch121}). 
In order to decrease the time spent for the model to learn the basic features, we self-trained a base VAE model on 300 TBP images. In the query time when a TBP image is given, we fine-tune the base model on the lesion images from the TBP image for a few epochs and then calculate reconstruction loss and features from lesion images. Using this approach, we were able to identify UDs in just a few seconds and get results as accurate as before.\\

Additionally, we observed that by incorporating a $\beta$-VAE's disentangled latent space and calculating the distance between each lesion's extracted features and mean of features belonging to the lesions in the same TBP image, we still can identify UDs and optionally can skip the finetuning phase.

\section{Data}
We evaluated our UD detection algorithm on 75 TBP images in total. 32 images were sourced from the SD-198 \cite{SD-198}, and SD-260 datasets \cite{SD-260}, and an additional 43 were collected internally from various clinics. UD lesions in these images were labelled by a board certified dermatologist. 
TBP images used in this experiment contained a varying number of lesions ranging from 10 to 182. Overall, we extracted a total of 4628 lesion images across all 75 TBP images. 53 of the images contained at least one lesion labelled as UD, with an average of 1.44 UDs on each TBP image. The validity of data was manually checked after the segmentation module.

\begin{figure*}[htp]
    \centering
    \includegraphics[width=\linewidth]{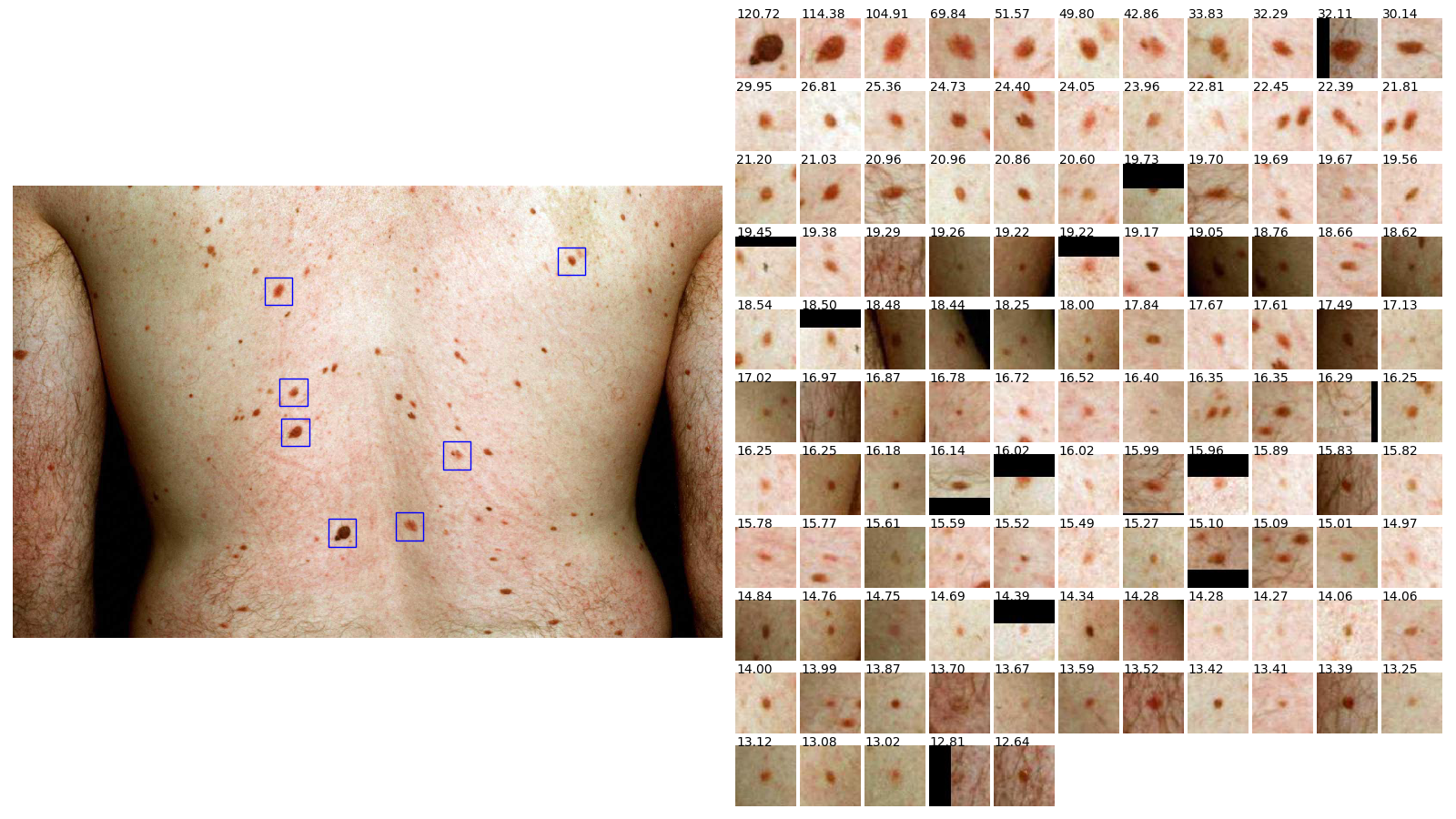}
    \caption{Automatic analysis of all the lesions in the image, with an ordered list of all the distance rankings, highest at top left hand side. The outliers which had a distance higher than our defined threshold of 38.63, are highlighted in blue on the image. The mean and standard deviation of the distances were 21.60 and 17.03 respectively.}
    \label{fig:UDresults}
\end{figure*}

\section{Results}
The UD detection problem can be considered either as ranking prediction problem or binary classification problem. In order to evaluate our algorithm's performance, we used ranking evaluation metrics and binary classification evaluation metrics. In order to find the rankings, first we calculated an embedding for each lesion and then calculated the L2 distance between each embedding to the mean of embeddings. By sorting these distances, we generated the rankings. In the binary classification setting, we defined a threshold on each lesion in order to determine whether a lesion should be called UD or not. Using the following formula, we obtained the threshold for each TBP image:
\begin{equation}
    threshold = mean(distances) + \min
  \begin{cases}
    mean(distances)\\
    std(distances)
  \end{cases}
\end{equation}

Fig \ref{fig:UDresults} shows the results of automated analysis of all the lesions on the image in Fig \ref{fig:backHB}. The 7 lesions outlined in blue rectangles are the outliers according to our threshold.

\subsection{Ranking Evaluation Metrics}
We used average precision (AP) and reciprocal rank (RR) as our primary metrics. Additionally we used top-3 and top-7 agreement (Agr.) from \cite{SoenksenUD2021}. We calculated each metric for all TBP images and report the average over all TBP images. In order to calculate top-3 agreement, for each lesion labelled as UD, we found its rank in our results. If the ranking was less than or equal to 3, we called success (TBP image counted as 1), otherwise we called it failure (TBP image counted as 0). We calculate and report the average of success over TBP images. Top-7 agreement is also calculated in a similar manner. Since not every TBP image contains a UD lesion, we only calculate top-3 and top-7 agreement for the TBP images containing at least one lesion labelled as UD.\\
Results are shown in  Table \ref{table:ranking_metrics}. 

\begin{table}[h!]
\centering
\begin{tabular}{| c | c | c | c |} 
 \hline
 MAP & MRR & Top-3 Agr. & Top-7 Agr. \\ [0.5ex] 
 \hline
 0.659 & 0.721 & 86.79\% & 94.34\% \\ [0.5ex]
 \hline
\end{tabular}
\caption{UD Ranking Evaluation}
\label{table:ranking_metrics}
\end{table}

\subsection{Binary Classification Metrics}
Binary classification predictions can be obtained by applying a threshold on the distances calculated for each lesion.Results are shown in Table \ref{table:binary_classification_metrics}. On average our model predicts $4.25$ UDs per TBP image. In order to evaluate our method, we measured accuracy, sensitivity, and specificity over all extracted lesions and also averaged per TBP image.

\begin{table*}[h!]
\centering
\begin{tabular}{| c | c | c |} 
 \hline
 \backslashbox{Metric}{Evaluation Type} & Over All Lesions (Micro Avg.) & Averaged Over TBPs (Macro Avg.) \\ [0.5ex] 
 \hline
 Accuracy & 94.16\% & 94.23\% \\ [0.5ex]
 \hline
 Sensitivity & 72.07\% & 71.91\% \\ [0.5ex]
 \hline
 Specificity & 94.70\% & 94.95\% \\ [0.5ex]
 \hline
\end{tabular}
\caption{UD Binary Classification Evaluation. It should be noted that when calculating average sensitivity over TBP images, only images with at least one UD lesion were considered.}
\label{table:binary_classification_metrics}
\end{table*}

\begin{figure}[h!]
\centering
\begin{subfigure}{0.49\linewidth}
\includegraphics[width=\linewidth]{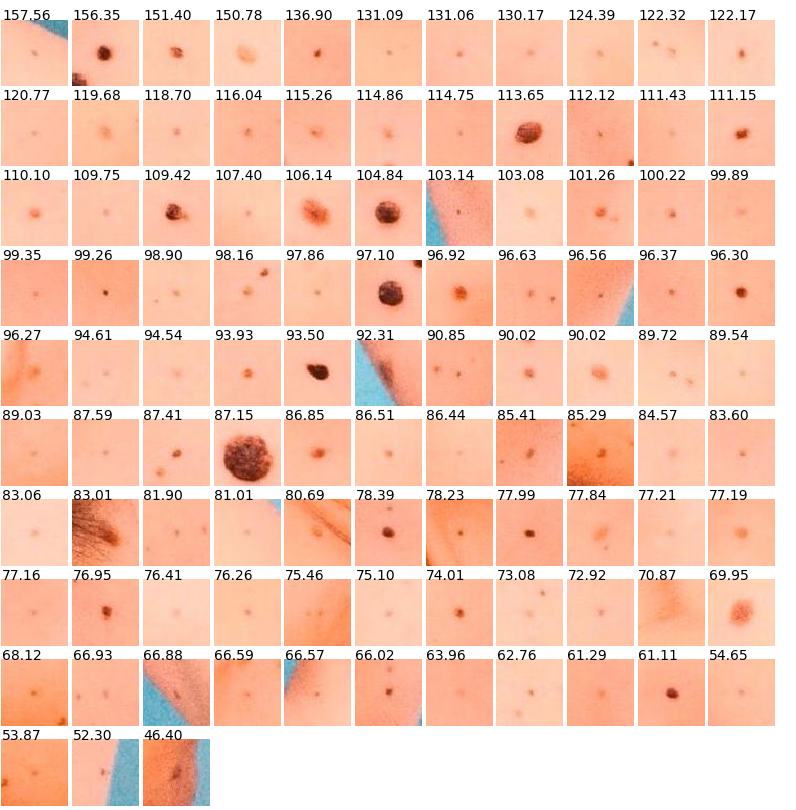} 
\caption{Before segmentation}
\label{fig:before_segmentation}
\end{subfigure}
\begin{subfigure}{0.49\linewidth}
\includegraphics[width=\linewidth]{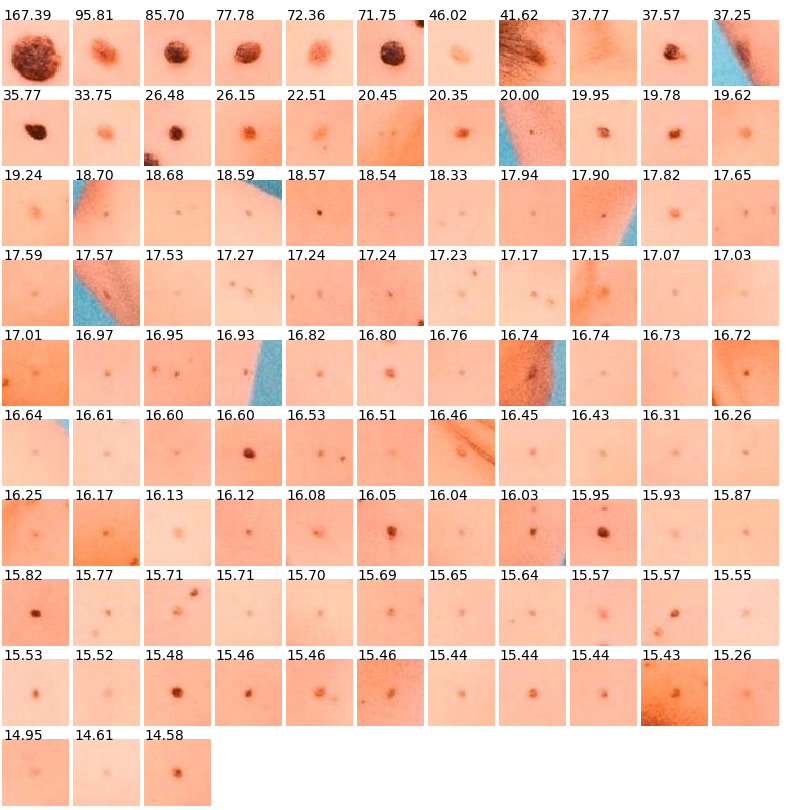}
\caption{After segmentation}
\label{fig:after_segmentation}
\end{subfigure}
\caption{Ranking predicted before and after incorporating segmentation module. Following the same rules as \ref{fig:UDresults}}
\label{fig:SegmentationComparison}
\end{figure}

\section{Discussion}
It is sometimes difficult to distinguish pigmented nevi from other skin conditions, such as lentigos and ephelides (freckles) which often occur on sun-exposed areas of the upper back, face, back of hands, and forearms. Also, differentiating small nevi (those less than 2 mm in diameter) from ephelides is not easy even clinically.
Depending on the image resolution, our algorithm automatically detects all the skin lesions larger than around 1.5mm diameter, and ranks them in order of similarity to each other, providing an objective measure of the outlier nature of each lesion. 
Our algorithm provides an opportunity to perform quick and unbiased screening for ugly duckling lesions in TBP images, especially useful for patients with many lesions. Additionally, our method can work on all types of skin lesions, which means it is not limited to pigmented lesions. \\

Our algorithm also benefits much from incorporation of segmentation module. Before incorporating segmentation module our results were heavily influenced by noise. Some removed noises were shadow, skin color around a lesion, and proximity to the body edge. With the help of the segmentation module, our model made much better predictions as it became more robust to noise and avoided many false positives. Fig \ref{fig:SegmentationComparison} shows how our ranking was improved with the help of segmentation module. It is important to mention that segmentation module also resolves the need for additional normalization methods to fix the lighting. Although normalization methods are mostly helpful, but sometimes they can drastically decrease the quality of images in some edge-cases.\\

Limitations of our method include difficulties in detecting UD lesions when the skin around the lesion has useful information for making the decision (e.g. scars after excision), as we are not able to incorporate additional domain knowledge in our decision making. We also remove hairs during detection and segmentation, which means a loss of a clue such as a hair growing out of a lesion, which indicates it might have been a congenital nevus.
Since our model does not have domain knowledge, it can not leverage these facts to make better decisions. 
\section{Conclusion and Future Work}
We have developed an algorithm to quickly identify outlier lesions in total body photography images, at even the relatively low resolution obtained by imaging backs with mobile smartphone cameras. Higher resolution images of lesions on arms, etc. can be analysed with even more accuracy. This opens up the opportunity to use this as a screening aid for patients to submit their photographs using teledermatology.

Future work will involve further evaluation of the method against a larger image set of expertly labelled images. We may also use different distance measures other than the L2 to determine outliers. 
Also we hope the result of this UD outlier detection project can improve classification of skin lesions extracted from TBP images.
We hope introduction of our pipeline opens the path for future research in the domain of TBP images.

{\small
\bibliographystyle{ieee_fullname}
\bibliography{egbib}
}

\end{document}